%% file: _robotPoseData.tex
\documentclass[10pt,twocolumn,letterpaper]{article}
\usepackage{cvpr}      %
\usepackage{graphicx}
\usepackage{amsmath}
\usepackage{amssymb}
\usepackage{booktabs}
\usepackage[pagebackref,breaklinks,colorlinks]{hyperref}
\usepackage[accsupp]{axessibility}  %
\usepackage{tabularx}
\usepackage{color}

\definecolor{ben}{rgb}{0.9,0.,0.5}

\definecolor{pw}{rgb}{0.4,0.6,0.0}

\definecolor{hj}{rgb}{0.5,0.0,0.5}

\definecolor{todo}{rgb}{1.0, 0., 0.}

\usepackage{pifont}%
\newcommand{\cmark}{\ding{51}}%

\usepackage{textcomp} %

\usepackage[capitalize]{cleveref}
\crefname{section}{Sec.}{Secs.}
\Crefname{section}{Section}{Sections}
\Crefname{table}{Table}{Tables}
\crefname{table}{Tab.}{Tabs.}

\begin{document}

\title{PhoCaL: A Multi-Modal Dataset for Category-Level Object Pose Estimation\\with Photometrically Challenging Objects}

\author{
\hspace{-12pt}
Pengyuan Wang$^{\ast 1}$, 
HyunJun Jung$^{\ast 1}$, 
Yitong Li$^{1}$, 
Siyuan Shen$^{1}$,
Rahul Parthasarathy Srikanth$^{1}$,\\
Lorenzo Garattoni$^{2}$,
Sven Meier$^{2}$,
Nassir Navab$^{1}$,
Benjamin Busam$^{1}$\\
$^{\ast}$ Equal Contribution\qquad
$^1$ Technical University of Munich\qquad
$^2$ Toyota Motor Europe\\
{\tt\small pengyuan.wang@tum.de}
\quad {\tt\small hyunjun.jung@tum.de}
\quad {\tt\small b.busam@tum.de}
}

\twocolumn[{%
\renewcommand\twocolumn[1][]{#1}%
\maketitle
\begin{center}
    \captionsetup{type=figure}
    \includegraphics[width=\linewidth]{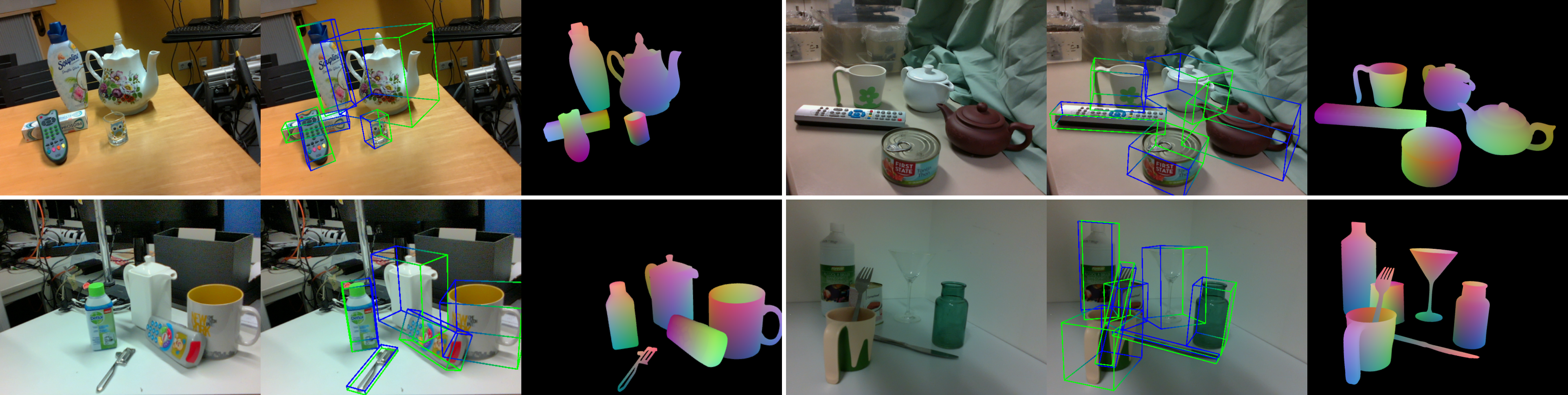}
    \captionof{figure}{PhoCaL comprises 60 high quality 3D models of household object in 8 categories with different photometric complexity. The selected objects include challenging texture-less, occluded, symmetric, reflective and transparent objects. Our robotic-induced pose annotation pipeline provides highly accurate 6D pose labels even for objects that are hard to capture by modern RGBD sensors. The figure shows RGB, 3D bounding boxes and rendered Normalized Object Coordinate Space (NOCS) map for 4 example scenes.
    }
    \label{fig:teaser}
\end{center}%
}]

\input{sections/00_abstract}

\input{sections/02_intro}

\input{sections/04_related_work}

\input{sections/06_dataset}
\input{sections/08_experiments}

\input{sections/10_conclusion}

{\small
\bibliographystyle{ieee_fullname}
\bibliography{literature}
}

\end{document}


\title{PhoCaL - Supplementary Material} %

\author{
\hspace{-12pt}
Pengyuan Wang$^{\ast 1}$, 
HyunJun Jung$^{\ast 1}$, 
Yitong Li$^{1}$, 
Siyuan Shen$^{1}$,
Rahul Parthasarathy Srikanth$^{1}$,\\
Lorenzo Garattoni$^{2}$,
Sven Meier$^{2}$,
Nassir Navab$^{1}$,
Benjamin Busam$^{1}$\\
$^{\ast}$ Equal Contribution\qquad
$^1$ Technical University of Munich\qquad
$^2$ Toyota Motor Europe\\
{\tt\small pengyuan.wang@tum.de}
\quad {\tt\small hyunjun.jung@tum.de}
\quad {\tt\small b.busam@tum.de}
}

\twocolumn[{%
\renewcommand\twocolumn[1][]{#1}%
\maketitle
\begin{center}
    \captionsetup{type=figure}
    \includegraphics[width=\linewidth]{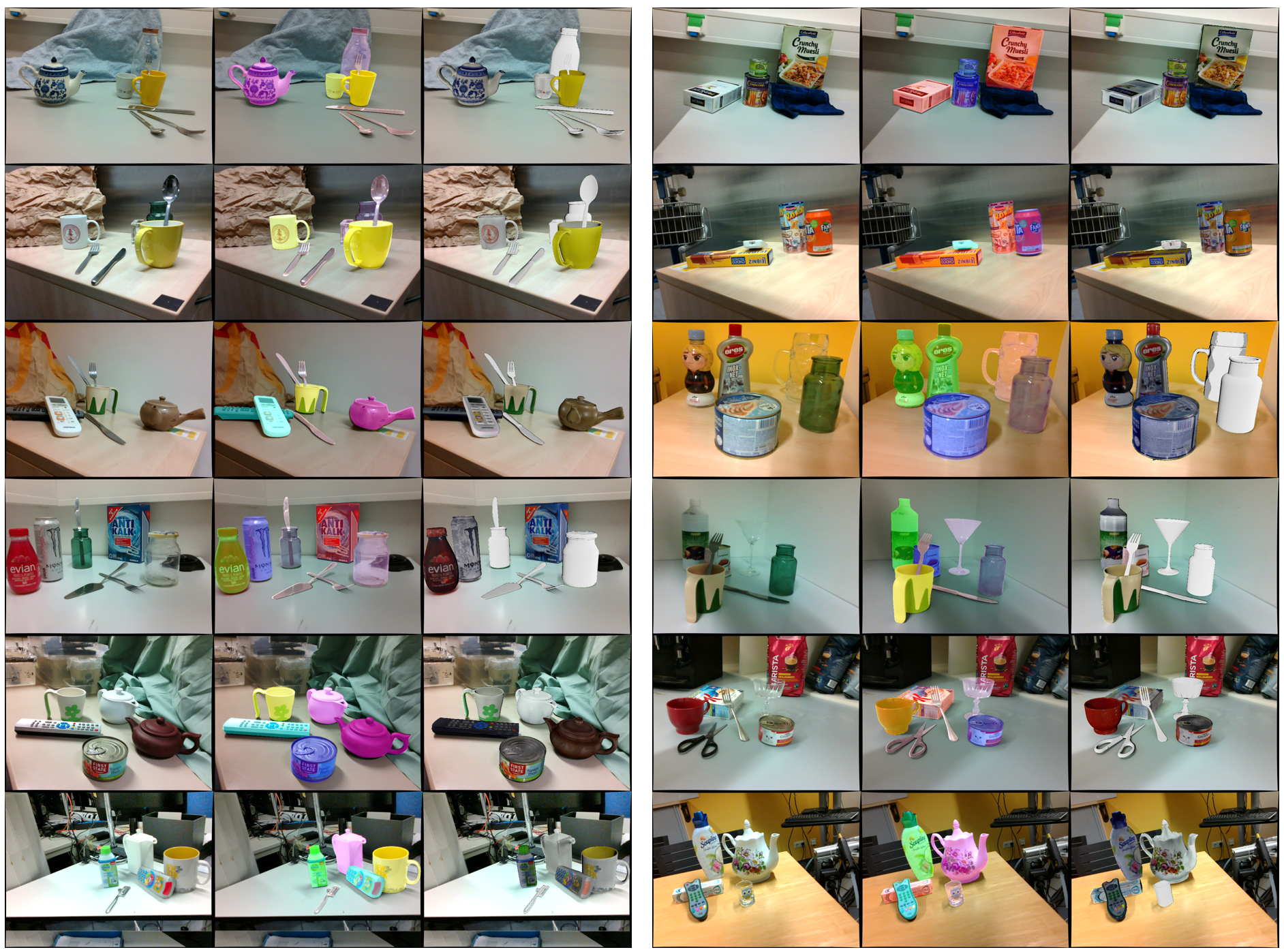}
    \captionof{figure}{Example images from all scenes in PhoCaL dataset. The figure shows RGB, coloured masks and rendered models in scenes. Note that the ground truth annotations are accurate even for photometrically challenging objects. }
    \label{fig:all_scenes}
\end{center}%
}]

\begin{figure*}[t]
 \centering
    \includegraphics[width=\linewidth]{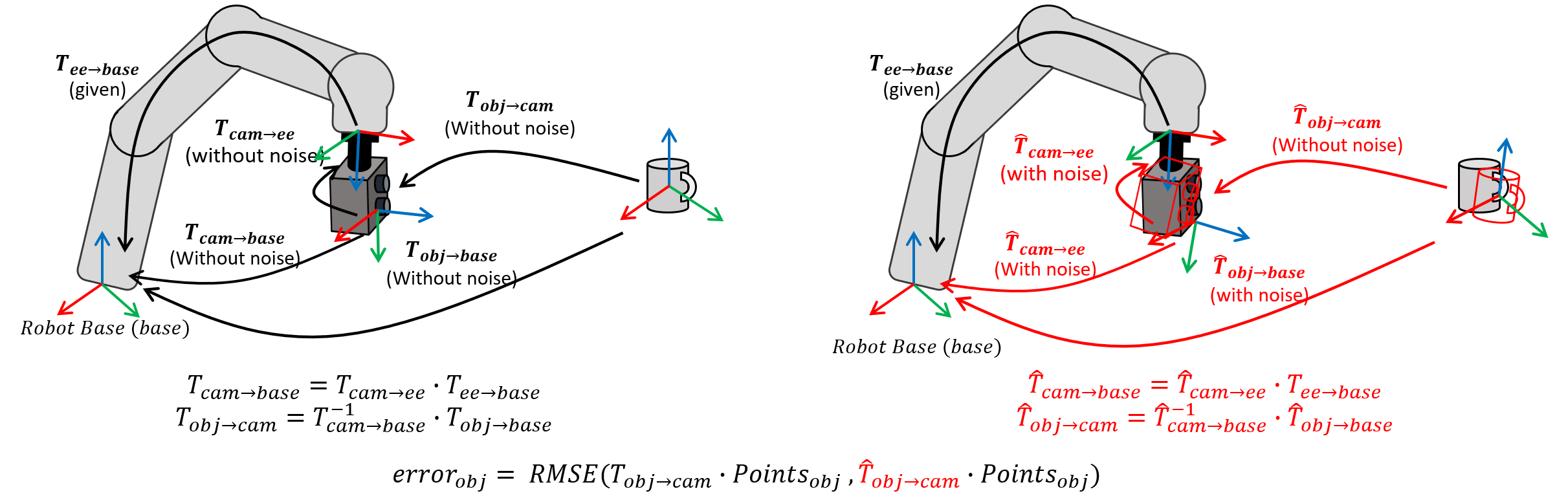}
    \caption{Pose graph for our simulated annotation evaluation setup. RMSE of pointwise error is calculated from object mesh points with pose from camera base with noise (\(\hat{T}_{obj->cam}\)) and without noise (\(T_{obj->cam}\)). }
    \label{fig:annotation_evaluation}
\end{figure*}

\section{Scene Example Visualization}
To make the dataset challenging and similar to real environments, different backgrounds are chosen with occlusion between objects. The detailed views of setups and backgrounds are visualized in Fig.~\ref{fig:all_scenes}. Our dataset is composed of 12 scenes with two different trajectories per each scene (i.e. a total of 24 trajectories).

\section{Details of Annotation Quality Evaluation}

To evaluate overall annotation quality of our dataset, we run simulated data acquisition with pre-calculated error statistics on object pose annotation step (Sec 3.4 in main paper) and hand-eye-calibration step (Sec 3.5 in main paper), then compare with ground truth similar to \cite{liu2020keypose, liu2021stereobj}. However, as our error statistics per step are obtained in 3D (translation) and 6D (translation + rotation), instead of a projection error in pixel as in \cite{liu2020keypose, liu2021stereobj}, the acquisition is simulated by directly applying the error statistics on the steps. In this section, we describe the details of the simulated evaluation pipeline.

\paragraph{Simulated Scene Setup}
To simulate the dataset acquisition in a realistic way, we chose scene 9 (Fig.~\ref{fig:all_scenes} 6th column, 2nd row) to evaluate our hand-eye calibration accuracy. All the objects are synthetically placed with their annotated pose from the robot base (\(T_{obj\to base}\)). Then the recorded trajectory of each camera is repeated by applying hand-eye calibration matrix (\(T_{cam\to ee}\)) on the end-effector pose (\(T_{ee\to base}\)). Here, the absolute ground truth pose of the objects from each camera center (\(T_{obj\to cam}\)) is obtained as follows (Fig.~\ref{fig:annotation_evaluation}, left):
\begin{equation}\label{eq:absolute_gt}
    T_{gt} = T_{obj\to cam} = T_{cam\to ee}^{-1} \cdot T_{ee\to base}^{-1} \cdot T_{obj\to base}
\end{equation}
The simulated annotated pose from the camera (\(\hat{T}_{obj\to cam}\)) is obtained by applying noise on both \(T_{obj\to base}\) and \(T_{cam\to ee}\), where we denote as \(\hat{T}_{obj\to base}\), \(\hat{T}_{cam\to ee}\) (Fig.~\ref{fig:annotation_evaluation}, right):
\begin{equation}\label{eq:annotated_gt}
    T_{annotated} = \hat{T}_{obj\to cam} = \hat{T}_{cam\to ee}^{-1} \cdot T_{ee\to base}^{-1} \cdot \hat{T}_{obj\to base}
\end{equation}

\paragraph{Simulated Error on Object Pose Annotation}
For each object in the scene, translation noise of \(0.20\,\text{mm}\) and rotation noise of $0.38^{\circ}$ (Sec 3.4 in the main paper) is added on the \(T_{obj\to base}\). To add randomness, we first generate two 3D unit vectors with random orientation per object, where the first vector is multiplied  by \(0.20\,\text{mm}\) for the translation error \(t_{error}\) and the second vector is utilized  as axis in axis-angle representation with an angle of $0.38^{\circ}$, for the rotation error \(R_{error}\).
\begin{equation}\label{eq:obj_error}
    \hat{T}_{obj\to base} = [R_{error}|t_{error}] \cdot T_{obj\to base}
\end{equation}

\paragraph{Simulated Error on Hand-Eye Calibration}
To add noise on the hand-eye calibration matrix, a small perturbation is applied on each camera's hand-eye calibration matrix. We apply random perturbation multiple times on the matrix, and choose the perturbation which gives error range of \(\text{RMSE}_{\text{RGBD}} = 0.89 \, \text{mm}\) and  \(\text{RMSE}_{\text{Polarization}} =  0.83 \, \text{mm}\) on \(\hat{T}_{cam\to ee}\) (Sec 3.5 in the main paper) as the simulated error on hand-eye calibration.

\paragraph{Simulated Error on Object Pose from Camera}
Two real trajectories of the end-effector poses \(T_{ee\to base}\) are used in the test. For each camera, we run the two sequences and calculate the pointwise error from each object's mesh obtained from \(T_{gt}\) and \(T_{annotated}\) (equation \ref{eq:absolute_gt} and \ref{eq:annotated_gt}). The RMSE error is calculated through the frames and averaged for each object. In the test, the final RMSE error is 0.84 mm for RGBD camera, and 0.76 mm for the polarization camera.

\begin{figure*}[t]
 \centering
    \includegraphics[width=\linewidth]{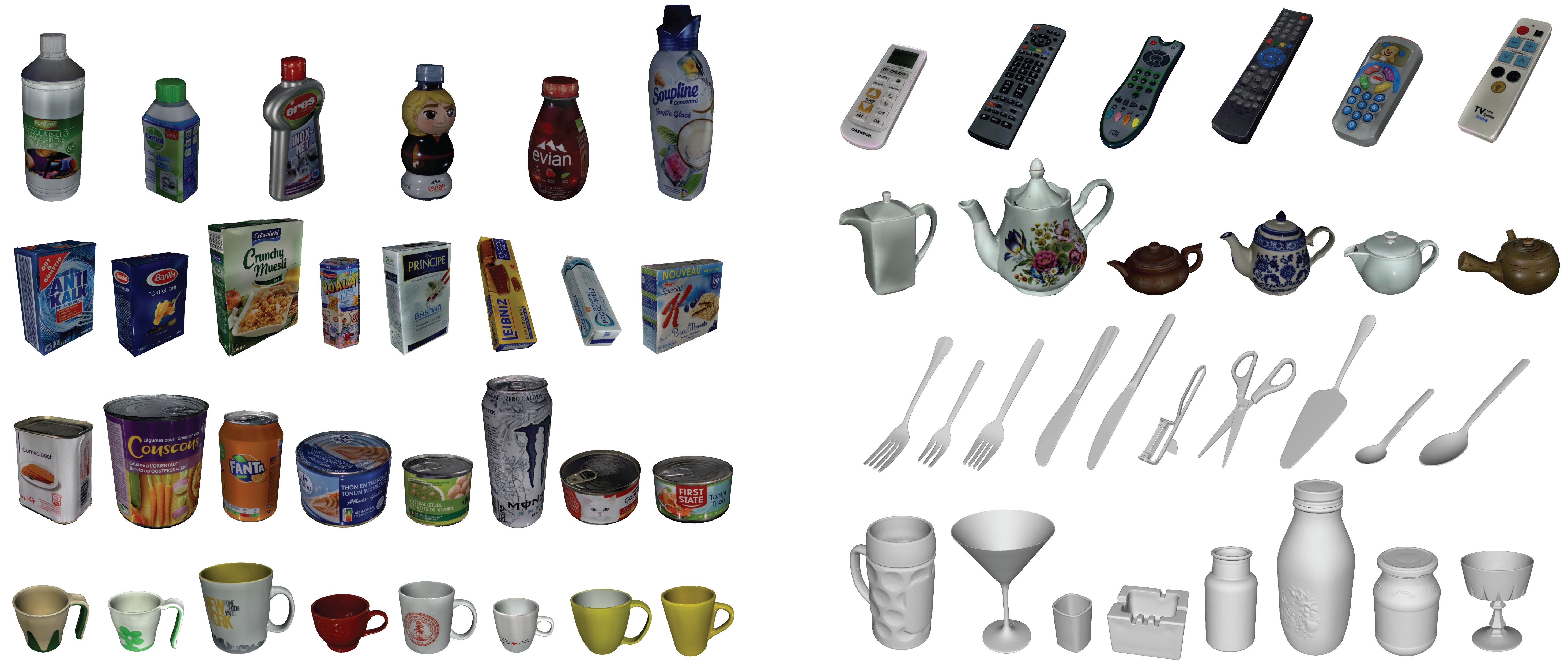}
    \caption{Illustration of high-quality 3D object models from all categories used in the dataset. On the left side are bottles, boxes, cans and cups. On the right side are remotes, teapots, cutlery and glassware.}
    \label{objects_models}
\end{figure*}

\begin{figure*}[t]
 \centering
    \begin{subfigure}[t]{0.5\textwidth}
        \centering
        \includegraphics[width=0.9\linewidth]{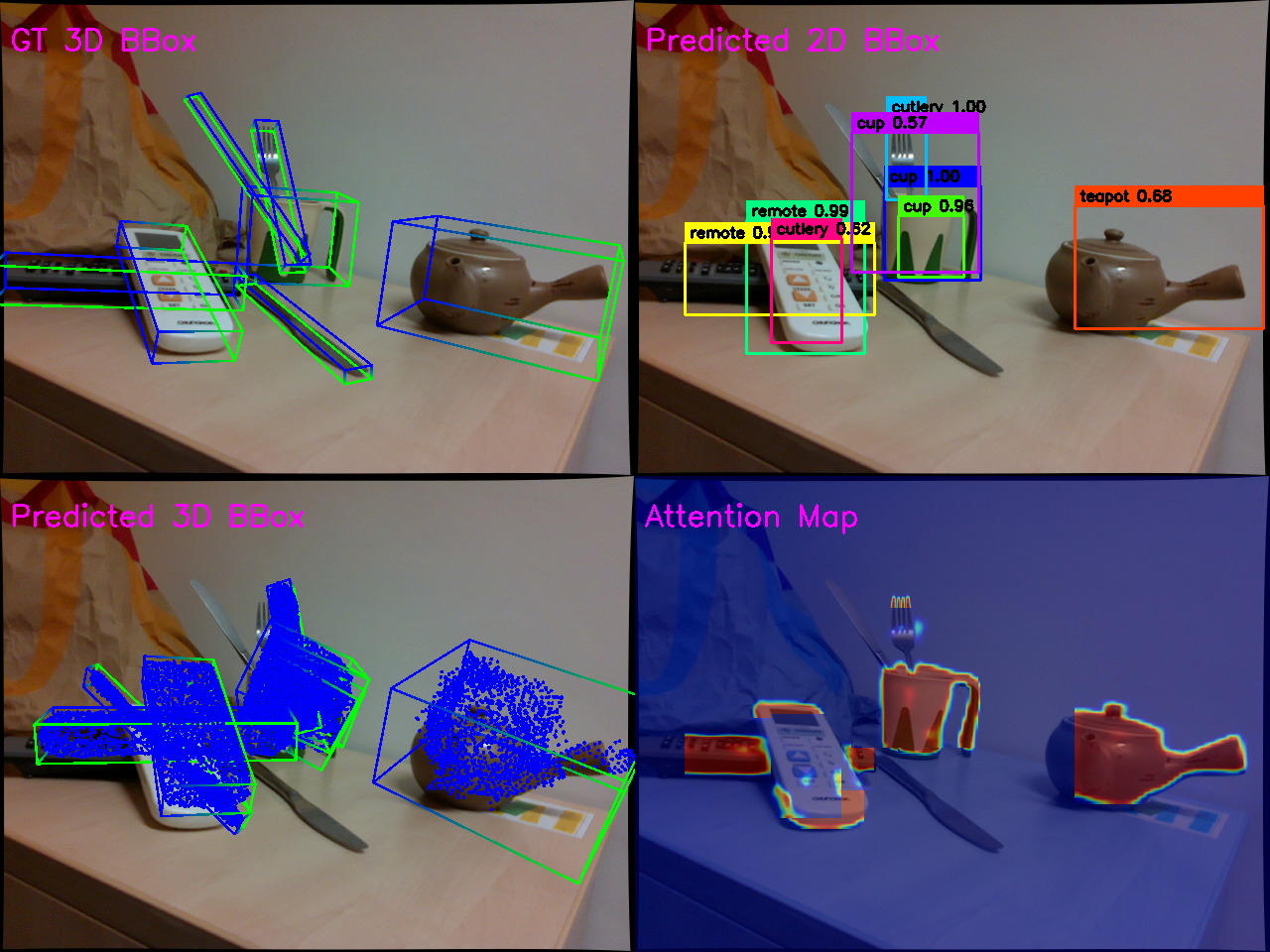}
    \end{subfigure}%
    \begin{subfigure}[t]{0.5\textwidth}
        \centering
        \includegraphics[width=0.9\linewidth]{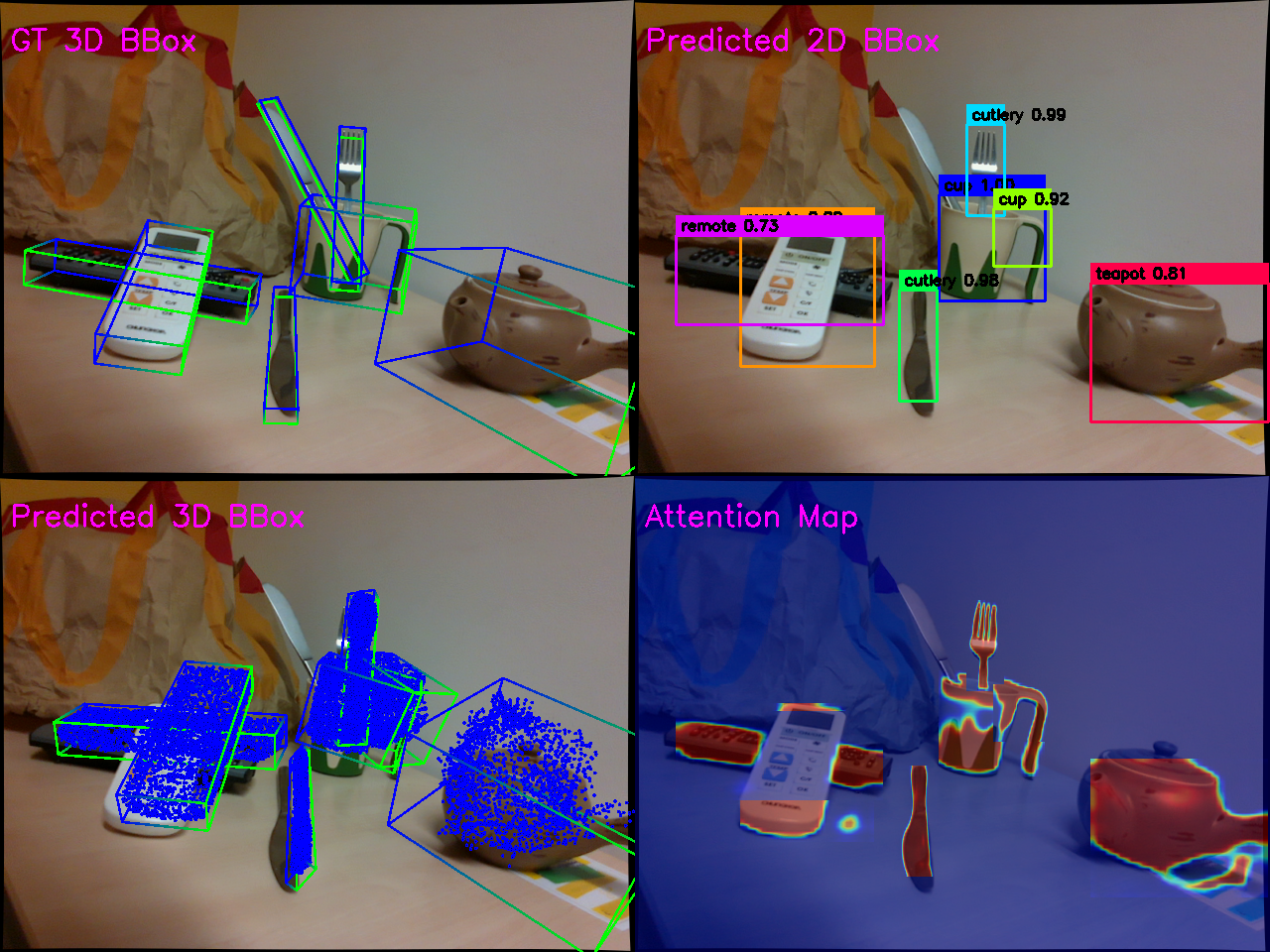}
    \end{subfigure}%
    \caption{Two example of CPS result for test images in experiment 1, in each example the result contains the ground truth 3D bounding boxes,  the predicted 2D bounding boxes, the predicted 3D bounding boxes, the attention maps}
    \label{cps_result}
\end{figure*}

\section{3D Object Models in the Dataset}

PhoCal comprises high quality 3D models for 60 objects in 8 categories. Textured models are available in 6 categories for bottles, boxes, cans, cups, remotes, and teapots. Photometrically very challenging objects without texture are given in 2 categories, namely cutlery (which are highly reflective) and glassware (which are transparent). The high quality models are visualized in Fig. \ref{objects_models}.

\section{Visualization of CPS Result}
The testing results of CPS in experiment 1 is visualized in Fig. \ref{cps_result}, where ground truth 3D bounding boxes, predicted 2D bounding boxes, predicted 3D bounding boxes, and attention maps are plotted. More visualizations are included in the supplementary video.

{\small
\bibliographystyle{ieee_fullname}
\bibliography{literature}
}

%% file: sections/00_abstract.tex
\begin{abstract}
Object pose estimation is crucial for robotic applications and augmented reality. Beyond instance level 6D object pose estimation methods, estimating category-level pose and shape has become a promising trend. As such, a new research field needs to be supported by well-designed datasets. To provide a benchmark with high-quality ground truth annotations to the community, we introduce a multimodal dataset for category-level object pose estimation with photometrically challenging objects termed PhoCaL. PhoCaL comprises 60 high quality 3D models of household objects over 8 categories including highly reflective, transparent and symmetric objects. 
We developed a novel robot-supported multi-modal (RGB, depth, polarisation) data acquisition and annotation process. It ensures sub-millimeter accuracy of the pose for opaque textured, shiny and transparent objects, no motion blur and perfect camera synchronisation. 

To set a benchmark for our dataset, state-of-the-art RGB-D and monocular RGB methods are evaluated on the challenging scenes of PhoCaL.
\end{abstract}

%% file: sections/02_intro.tex
\begin{figure*}[htbp]
 \centering
    \includegraphics[width=\linewidth]{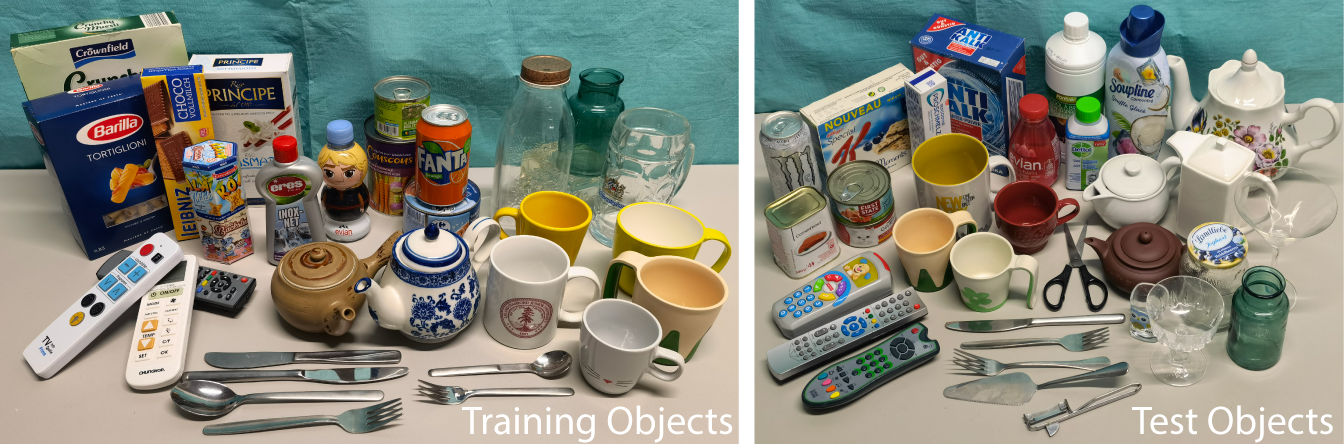}
    \caption{Our dataset comprises 60 household objects among 8 object categories. The training and test split is depicted here.}
    \label{dataset_split}
\end{figure*}
\section{Introduction}
\label{sec:intro}

Vision systems interacting with their environment need to estimate the position and orientation of objects in space, which highlights why 6D object pose estimation is an important task for robotic applications. Even though there have been great advances in the field \cite{bukschat2020efficientpose,zakharov2019dpod}, instance-level 6D pose methods require pre-scanned object models and support limited number of objects. Category-level object pose estimation ~\cite{wang2019normalized} scales better to the needs of real operating environments. However, photometrically challenging objects such as shiny, e.g. metallic, and  transparent, e.g. glass, objects are very common in our daily life and little work has been done to estimate their 6D poses within practical accuracy on a category-level. The difficulty arises from two aspects: first, it is difficult to annotate 6D pose ground truth for photometrically challenging objects since no texture can be used to determine key points; second, commonly used depth sensors fail to return the correct depth information, as structured light and stereo method often fail to correctly interpret reflection and refraction artefacts. As a consequence, RGB-D methods~\cite{wang2019normalized,lin2021dualposenet} do not work reliably with photometrically challenging objects.
We introduce PhoCaL, a class-level dataset of photometrically challenging objects with high-quality ground-truth annotations. 
The dataset provides multi-modal data such as RGB, depth and polarization which enables investigation into object's surface reflectance properties. 

We obtain highly accurate ground truth poses with a novel method using a collaborative robot arm in gravity compensated mode and a calibrated mechanical tip. In order to annotate the 6D pose of transparent and non-textured objects, a specially designed tip is mounted on the robot arm.  With the calibrated tip, the positions of pre-defined points on the object surface are acquired on the real object and matched to a scan thereof. Using this method, the object pose can be determined with an order of magnitude more accuracy than previous methods. For transparent and textureless objects, topographic key points are used instead of textural ones. The points gathered in this way are then matched to the object model in a final ICP~\cite{besl1992method} step to yield an accurate fit.%

The camera to robot end-effector transformation is needed to obtain the object poses in camera coordinates. Typically, hand-eye calibration approaches solve this by visually estimating the marker position and optimizing for the transformation between camera and end-effector. To minimize the error propagation and obtain highly accurate ground truth labels, we instead used the end-effector tip of the arm in gravity-compensated mode to measure the position of 12 points on a ChArUco \cite{an2018charuco} board. This allows us to use the robot's accurate position system to obtain both object poses and camera poses for image sequences.

Beyond photometrically challenging categories and high-quality annotations, multi-modal input is another highlight of PhoCaL. As the active depth sensors fail on metallic and transparent surfaces, we include an additional passive sensor modality in the form of a polarization camera. It provides valuable information on object surfaces~\cite{kalra2020deep}. In our setup, we designed and 3D printed a rig that holds multiple cameras, each mounted on it and carefully calibrated. During recording, a pre-defined trajectory is repeated by the robot arm. The robot arm stops when capturing images from all cameras, which avoids motion blur and diminished effects from imperfect synchronization.

In summary, our main contributions are:
\begin{enumerate}
    \item We propose \textbf{PhoCaL}, a \textbf{multi-modal} (RGBD + RGBP) \textbf{dataset for category-level object pose estimation}. The dataset comprises 60 high-quality 3D models of household objects including symmetric, transparent and reflective objects in 8 categories with 24 sequences featuring occlusion, partial visibility and clutter.
    \item We introduce a new and \textbf{highly accurate pose annotation method using a robotic manipulator} that allows for sub-millimeter precision 6D pose annotations of photometrically challenging objects even with reflective or transparent surfaces.
\end{enumerate}

%% file: sections/04_related_work.tex
\begin{table*}
\begin{tabularx}{\textwidth}{r | c c c | c c c | c c c c | r r r | l}
\toprule
  \small Dataset &
  \rotatebox[origin=c]{90}{\small RGB}          &
  \rotatebox[origin=c]{90}{\small Depth}        &
  \rotatebox[origin=c]{90}{\small Polarisation} &
  \rotatebox[origin=c]{90}{\small Real}         &
  \rotatebox[origin=c]{90}{\small Multi-View}   &
  \rotatebox[origin=c]{90}{\small Robotic GT}   &
  \rotatebox[origin=c]{90}{\small Occlusion}    &
  \rotatebox[origin=c]{90}{\small Symmetry}     &
  \rotatebox[origin=c]{90}{\small Transparent}  &
  \rotatebox[origin=c]{90}{\small Reflective}   &
  \rotatebox[origin=c]{90}{\small Categories}   &
  \rotatebox[origin=c]{90}{\small Objects}      &
  \rotatebox[origin=c]{90}{\small Sequences}       &
  \rotatebox[origin=c]{90}{\small License}      \\ 
\midrule
FAT~\cite{tremblay2018falling}                            & \cmark & \cmark &        &          & \cmark &        & \cmark & \cmark &        &          & --     & $21$   & $>1$k    & CC BY-NC-SA 4.0 \\
BlenderProc~\cite{denninger2019blenderproc}               & \cmark & \cmark &        &          & \cmark &        & \cmark & \cmark &        &          & --     & --     & $>1$k    & GNU GPL 3.0 \\
LabelFusion~\cite{marion2018label}                        & \cmark & \cmark &        & \cmark   &        &        & \cmark &        &        &          & --     & $12$   & $138$    & BSD 3-Clause \\
Toyota Light~\cite{hodan2018bop}                          & \cmark & \cmark &        & \cmark   &        &        &        & \cmark &        &          & --     & $21$   & $21$     & MIT \\
YCB~\cite{calli2015benchmarking,xiang2018posecnn}         & \cmark & \cmark &        & \cmark   &        &        & \cmark & \cmark &        &          & --     & $21$   & $92$     & MIT \\
Linemod~\cite{hinterstoisser2011multimodal,brachmann2014} & \cmark & \cmark &        & \cmark   &        &        & \cmark & \cmark &        &          & --     & $15$   & $15$     & CC BY 4.0 \\
GraspNet-1Billion~\cite{fang2020graspnet}                 & \cmark & \cmark &        & \cmark   &        &        & \cmark & \cmark &        &          & --     & $88$   & $190$    & CC BY-NC-SA 4.0 \\
T-LESS~\cite{hodan2017t}                                  & \cmark & \cmark &        & \cmark   &        &        & \cmark & \cmark &        &          & --     & $30$   & $20$     & CC BY 4.0 \\
HomebrewedDB~\cite{kaskman2019homebreweddb}               & \cmark & \cmark &        & \cmark   &        &        & \cmark & \cmark &        &          & --     & $33$   & $13$     & CC0 1.0 Universal \\
ITODD~\cite{Drost_2017_ICCV}                              &        & \cmark &        & \cmark   & \cmark &        & \cmark & \cmark &        & (\cmark) & --     & $28$   & $800$    & CC BY-NC-SA 4.0 \\
StereoOBJ-1M~\cite{liu2021stereobj}                       & \cmark &        &        & \cmark   & \cmark &        & \cmark & \cmark & \cmark & \cmark   & --     & $18$   & $183$    & Not (yet) released \\
\midrule
kPAM~\cite{manuelli2019kpam}                              & \cmark & \cmark &        & \cmark   &        &        & \cmark & \cmark &        &          & $2$    & $91$   & $362$    & MIT \\
CAMERA25~\cite{wang2019normalized}                        & \cmark & \cmark &        & (\cmark) &        &        & \cmark & \cmark &        &          & $6$    & $42$   & $30$     & MIT \\
REAL275~\cite{wang2019normalized}                         & \cmark & \cmark &        & \cmark   &        &        &        & \cmark &        &          & $6$    & $42$   & $13$     & MIT \\
TOD~\cite{liu2020keypose}                                 & \cmark & \cmark &        & \cmark   & \cmark &        &        & \cmark & \cmark &         & $3$    & $20$   & $10$     & CC BY 4.0 \\
Ours (PhoCaL)                                             & \cmark & \cmark & \cmark & \cmark   & \cmark & \cmark & \cmark & \cmark & \cmark & \cmark   & $8$    & $60$   & $24$     & CC BY 4.0
\end{tabularx}
{
\caption{Overview of pose estimation datasets. The upper part shows instance-level datasets while the lower part includes category-level setups. PhoCaL is the only dataset that includes both photometrically challenging objects with high quality (robotic) pose annotations and all three modalities, RGB, depth, and polarisation.}%
\label{tab:dataset_comparison}%
}
\end{table*}

\section{Related Work \& Current Challenges}
Standardized datasets are used in the field of object pose and shape estimation to quantify and compare contributions and advances in the field. These datasets generally fall in two domains: instance-level datasets, where the 3D model of the object is known a priori; and category-level datasets, where the exact CAD model is unknown. Tab.~\ref{tab:dataset_comparison} provides an overview of related datasets in both domains.

\subsection{Instance-level 6D Object Pose Dataset}
One of the earliest, most widely used publicly available datasets for instance level pose estimation is LineMOD~\cite{hinterstoisser2011multimodal} and its occlusion extension LM-Occlusion~\cite{brachmann2014}.
Their data was acquired using a PrimeSense RGB-D Carmine sensor and a marker board was used to keep track of the relative sensor pose. While undoubtedly pioneering this field, the 3D model quality is now outdated and the leader boards on these datasets have become saturated. HomebrewedDB~\cite{kaskman2019homebreweddb} accounts for the latter shortcoming by providing high quality 3D models scanned with a structured light sensor. Including three models from LineMOD, they add 30 more toy, household and industrial objects. Different illumination conditions and occlusions make the scenes more challenging. Other datasets also include household objects~\cite{rennie2016dataset, doumanoglou2016recovering, tejani2014latent, hodan2018bop} or focus on industrial parts~\cite{hodan2017t, Drost_2017_ICCV} with low texture for which it is also possible to manually design or retrieve accurate CAD models~\cite{hodan2017t}. The BOP 6D pose benchmark~\cite{hodan2018bop} includes a summary of these datasets with standardized metrics in a common format. 

While the datasets mentioned so far provide individual frames, the YCB-Video dataset~\cite{xiang2018posecnn} also includes video sequences of 21 household objects. While YCB uses LabelFusion~\cite{marion2018label} for semi-manual frame annotation and pose propagation through the sequence, Garon et al.~\cite{garon2018framework} leverage tiny markers on the object to estimate the poses in their videos directly at the cost of synthetic data cleaning afterwards. The advent of photo realistic rendering further enables a branch of works that leverages training on purely synthetic data~\cite{tremblay2018falling,denninger2019blenderproc}. Although this circumvents the cumbersome pose labelling process, it introduces a domain gap between synthetic data for evaluations and real-world appearances faced in the final applications.

\subsection{Category-level Object Poses and Datasets}
In real-world applications, a 3D model is not always available, but pose information is still required. Detection of such objects under these conditions has classically been tackled using 3D geometric primitives~\cite{carr2012monocular, birdal2018minimalist, birdal2019generic}.

While these methods consider outdoor scenes for which kitti~\cite{geiger2012we} provides 3D bounding box annotations, they lack object shape comparison and the information is often too inaccurate for robotic grasping tasks. The pioneering work of NOCS~\cite{wang2019normalized} was the first category-level method that could detect object pose and shape in indoor environments. %
Further investigations consider correspondence-free methods~\cite{chen2020learning} where a deep generative model learns a canonical shape space from RGBD and a method to estimate pose and shape for fully unseen objects is also proposed~\cite{park2020latentfusion}, albeit this method requires a reference image for latent code generation. CPS~\cite{manhardt2020cps} demonstrates how to estimate pose and metric shapes at category level, using only a monocular view. The extension CPS++~\cite{manhardt2020cps++} further utilizes synthetic data and a domain transfer approach using self-supervised refinement with a differentiable renderer from RGBD data without annotations. SGPA \cite{chen2021sgpa} explores shape priors to estimate the object pose. DualPoseNet \cite{lin2021dualposenet} leverages spherical fusion for better encoding of the object information. 

We leverage the standard RGBD method NOCS and the strong state-of-the-art RGB method CPS to set the baselines on our new dataset. While task-specific datasets for general object detection exist for robot grasping~\cite{manuelli2019kpam, fang2020graspnet}, methods for category-level pose estimation are typically tested on NOCS~\cite{wang2019normalized} data. The NOCS objects comprise various categories, but do not contain photometric challenges often present in everyday objects such as reflectance and transparency.

\subsection{Photometric Challenges and Multimodalities}
While texture-less objects~\cite{hodan2017t} were initially challenging for pose estimation, transparency presents an even bigger hurdle. While the problem is not new, previous methods have addressed this using RGB stereo without a 3D model to identify grasping points only~\cite{saxena2008robotic}. Rotational object symmetry can be leveraged by contour fitting for transparent object reconstruction~\cite{phillips2016seeing} using template matching.
ClearGrasp~\cite{sajjan2020clear} proposes a method for geometry estimation of transparent objects based on RGBD. However, this method passes over the transparent regions from the depth map and predicts depth from RGB in these areas instead.
Liu et al.~\cite{liu2020keypose} investigate instance- and category-level pose estimation from stereo imagery. Since their depth sensing fails on transparent objects, they use an opaque object twin as proxy to establish ground truth depth. More recently StereOBJ-1M proposed~\cite{liu2021stereobj} a large dataset including transparent and translucent objects with specular reflections and symmetry. However, at the time of this writing it is not yet available for download.

For 2D object detection, information from multiple orthogonal sensor modalities such as polarisation (RGBP) can help for transparent object segmentation~\cite{kalra2020deep}. This modality can provide information in regions were depth sensors fail. Their inherent connection with surface normals~\cite{zou20203d} can also make them attractive for pose estimation of photometrically challenging objects.

\begin{figure}[!t]
 \centering
    \includegraphics[width=\linewidth]{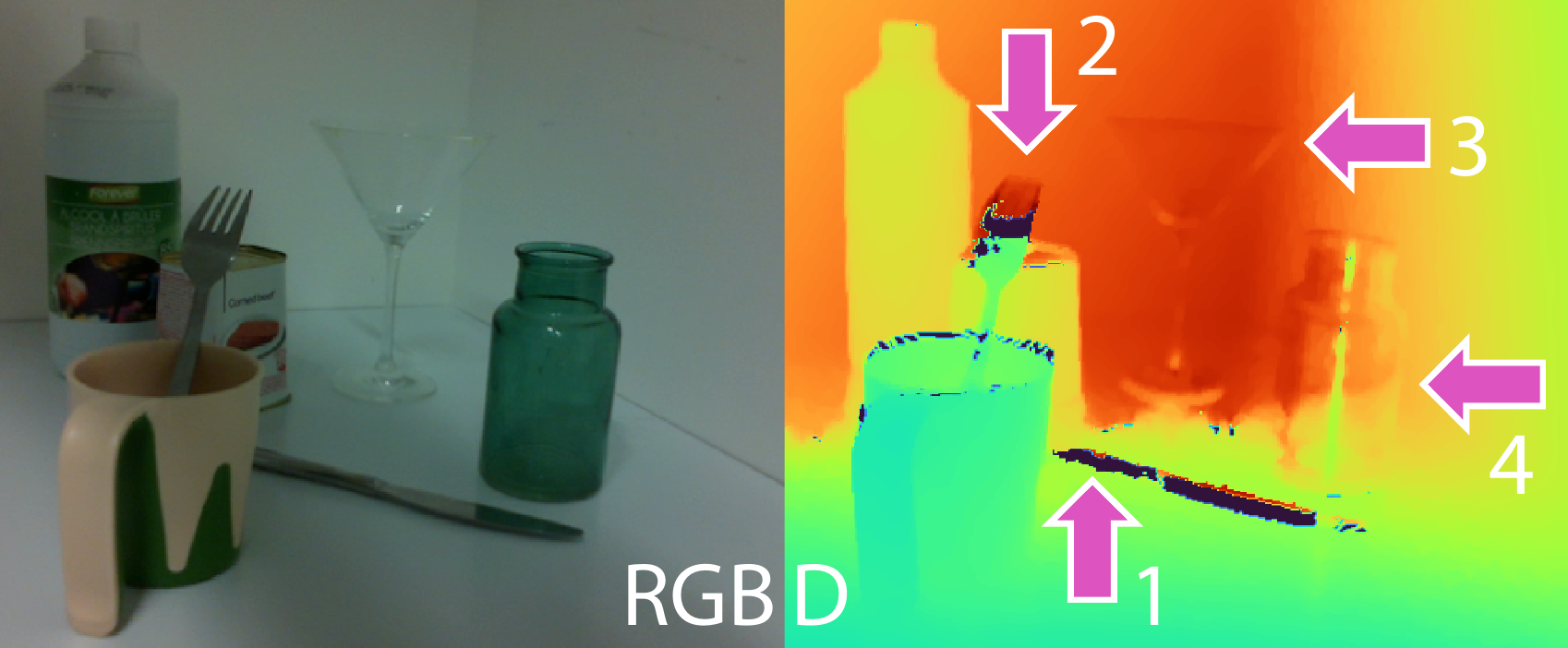}
    \caption{Limitations of RGBD sensors. The depth for photometrically challenging objects is difficult to measure with a commodity depth sensor. The intel RealSense D515 LiDAR ToF sensor used here is affected by reflections that lead to invalid (1) incorrect (2) distance estimates. Moreover, the glassware becomes invisible to the sensor (3) and causes noise (4).
    }
    \label{fig:rgbd_limits}
\end{figure}

\subsection{Ground Truth Pose Annotation}\label{subsec:annotation_problem}
Manual annotation of 6D pose is difficult and extremely time-consuming. Therefore, most datasets rely on semi-manual processes for ground truth annotation. The data from a depth sensor, if available, is often used to register the 3D model and manual adjustments are applied to visually refine the pose for this one frame. Relative camera motion is typically calculated using visual markers~\cite{hinterstoisser2011multimodal, kaskman2019homebreweddb} to propagate the pose information through a sequence of images. The use of depth sensors for ICP-based alignment of pose labels reduces labour and improves fully-manual annotation quality. However, depth maps from RGBD sensors are erroneous or invalid for photometrically challenging objects with high reflectance and translucent or transparent surfaces~\cite{liu2021stereobj}. An examples is shown in Fig.~\ref{fig:rgbd_limits}.

\begin{figure}[t!]
 \centering
    \includegraphics[width=\linewidth]{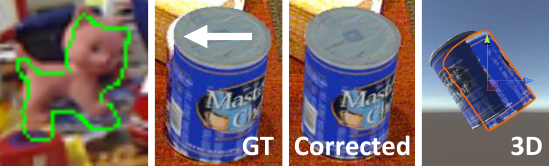}
    \caption{Annotation quality for poses in datasets Linemod~\cite{hinterstoisser2011multimodal} (projected green silhouette, left) and YCB~\cite{calli2015benchmarking} (rendered overlay, right) together with its correction~\cite{busam2020like} (right).}
    \label{fig:quality_others}
\end{figure}

Ensuring high quality of pose labels over a series of images is difficult and errors accumulate as the examples in Fig.~\ref{fig:quality_others} show. This equally affects depth-based refinement strategies of 6D pose pipelines~\cite{kehl2017ssd,hodan2018bop}. 
We propose a mechanical measurement process using a robotic manipulator to circumvent this issue and allow for high precision labels that omits the error propagation of relative camera pose retrieval from images.

%% file: sections/06_dataset.tex
\begin{figure*}[!t]
 \centering
    \includegraphics[width=\linewidth]{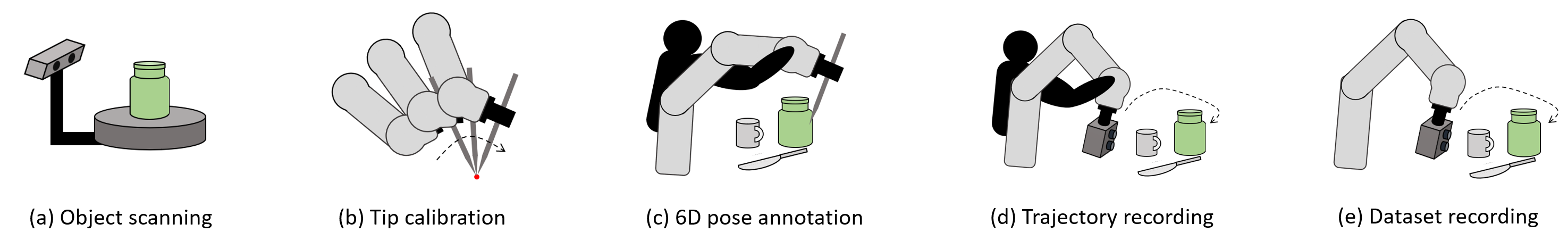}
    \caption{Overview of dataset acquisition pipeline. (a): 3D models are extracted with a structured light scanner. (b): Pivot calibration calibrates a tipping tool to robot coordinates. (c): 6D poses are annotated using the tool and manual movements of the robot. (d): The camera trajectory is saved. (e): Dataset is recorded automatically following the planned trajectory.}
    \label{fig:pipeline_overview}
\end{figure*}

\begin{figure*}[htbp]
 \centering
    \includegraphics[width=\linewidth]{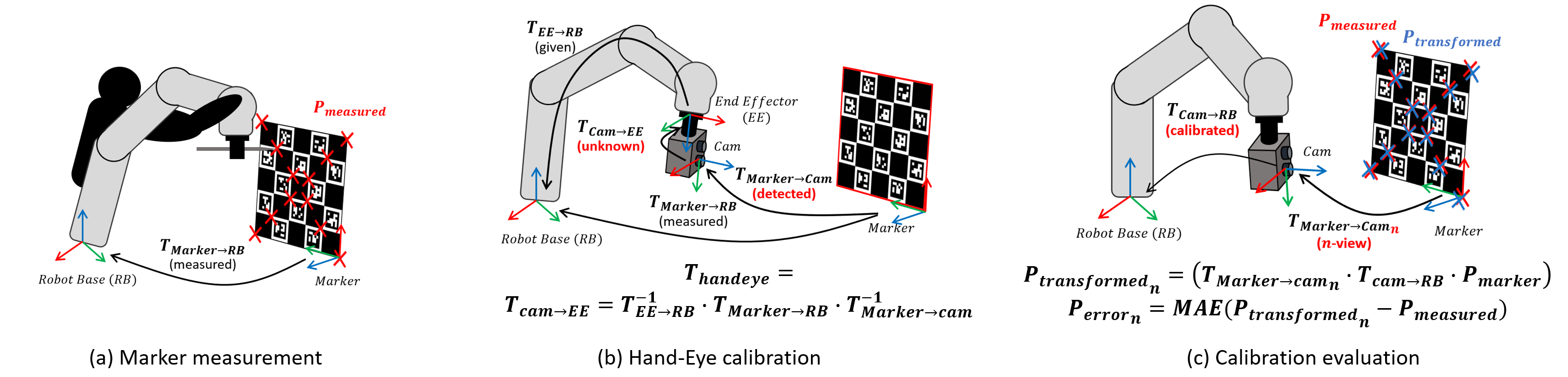}
    \caption{Overview of hand-eye-calibration and its evaluation. (a): shows the marker-to-robot calibration. (b): illustrates camera-to-robot hand-eye calibration. (c) depicts our accuracy evaluation.}
    \label{fig:handeye}
\end{figure*}

\section{Dataset Acquisition Pipeline}
Our dataset features multiple object classes including photometrically challenging classes such as objects with reflective surfaces or transparent material. It also provides multi-modal sensor data with highly accurate 6D pose annotation. This section describes our dataset acquisition pipeline as shown in Fig.~\ref{fig:pipeline_overview}.

\subsection{Objects Model Acquisition}
To represent a cross section of common household objects, we selected eight common categories for our category-level 6D object pose dataset: bottle, box, can, cup, remote, teapot, cutlery, glassware. All object models are scanned using an EinScan-SP 3D Scanner (SHINING 3D Tech. Co., Ltd., Hangzhou, China). The scanner is a structured light stereo system with a single shot accuracy of $\leq 0.05~\text{mm}$ in a scanning volume of $1200 \times 1200 \times 1200~\text{mm}^3$.

The models from the first six categories are provided as textured obj files. Since the cutlery and glassware objects are photometrically challenging with their highly reflective and transparent surfaces, we apply a self-vanishing 3D scanning spray (AESUB Blue, Aesub, Recklinghausen, Germany) to make the objects temporarily opaque for scanning. We scan the object and provide an obj file without texture. The spray sublimes after approx. $4$~h.

\subsection{Scene Acquistion Setup}
For each scene, 5-8 objects are placed on the table with the random background. We use a KUKA LBR iiwa 7 R800 (KUKA Roboter GmbH, Augsburg, Germany) 7 DoF robotic arm that guarantees a positional reproducibility of $\pm0.1$~mm.
The vision system comprises a Phoenix 5.0 MP Polarization camera (IMX264MZR/MYR) with Sony IMX264MYR CMOS (Color) Polarsens (i.e. PHX050S1-QC) (LUCID Vision Labs, Inc., Richmond B.C., Canada) with a Universe Compact lens with C-Mount 5MP 2/3'' 6mm f/2.0 (Universe, New York, USA).
As depth camera, the Time-of-Flight (ToF) sensor Intel\textsuperscript \textregistered RealSense\texttrademark LiDAR L515 is used, which acquires depth images at a resolution of 1024x768 pixels in an operating range between 25~cm and 9~m with a field-of-view of 70\textdegree x 55\textdegree and an accuracy of $5\pm2.5$~mm at $1$~m distance up to $14\pm15.5$~mm at $9$~m distance.

\subsection{Tip Calibration}
We use a rigid, pointy metallic tip to obtain the coordinate position of selected points on the object. Tip calibration is therefore essential to ensure the accuracy of the system. The rig attached to the robot's end-effector consists of custom 3D printed mount which holds the tool-tip rigidly. The pivot calibration is performed as shown in Fig.~\ref{fig:tipping} (left), where the tip point is placed in a fixed position, while only the robot end-effector position is changed. We collect data from N such tip positions with corresponding end-effector poses, ${}_{i}{T_e^b}$, which contain rotation ${}_{i}{R_e^b}$ and translation ${}_{i}{t_e^b}$, the final translation ${t_t^e}$ of the end-effector is calculated as follows:

\begin{equation}
    t_t^e = \begin{bmatrix}
    {}_{1}{R_e^b} - {}_{2}{R_e^b} \\
    {}_{2}{R_e^b} - {}_{3}{R_e^b} \\
    \vdots \\
    {}_{n}{R_e^b} - {}_{1}{R_e^b} \\
    \end{bmatrix}
    ^{\dagger} \cdot 
    \begin{bmatrix}
    {}_{1}{t_e^b} - {}_{2}{t_e^b} \\
    {}_{2}{t_e^b} - {}_{3}{t_e^b} \\
    \vdots \\
    {}_{n}{t_e^b} - {}_{1}{t_e^b} \\
    \end{bmatrix}
\end{equation}
where ${}^{\dagger}$ denotes the pseudo-inverse.
We evaluate the tip calibration by calculating the variance of each tip location at the pivot point. The variance of the tip location in our setup is $\varepsilon = 0.057~\text{mm}$.

\begin{figure*}[t!]
 \centering
    \includegraphics[width=\linewidth]{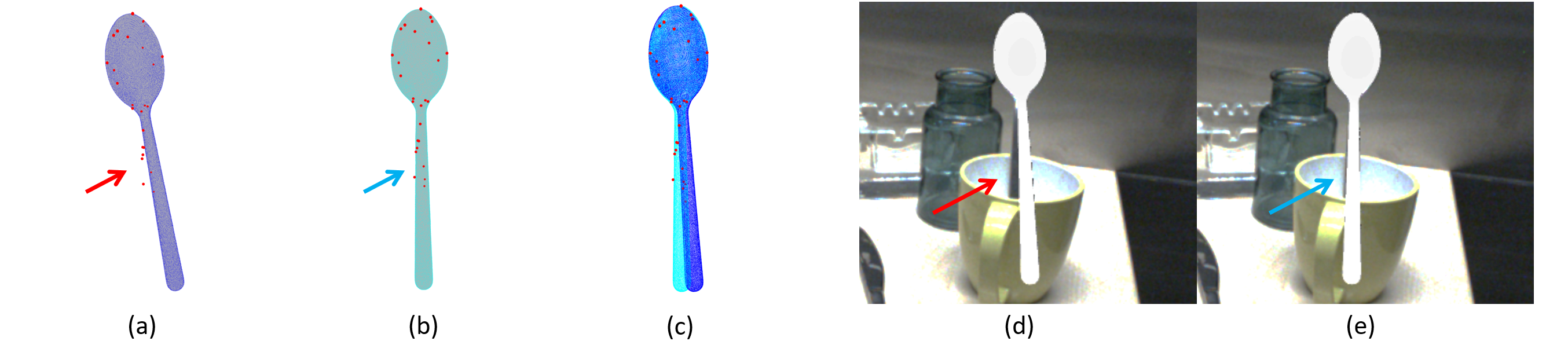}
    \caption{Example of annotation quality before and after ICP based refinement on the textureless object. (a) Initial pose of mesh overlaid with measured surface points (red dots) shows error in initial pose (red arrow). (b) After the ICP, refined pose matches with the surface points properly (blue arrow). (c) Shows improvement in 6D pose annotation. Rendering of the mesh with initial pose (d) and refined pose (e) shows a significant difference in quality.}
    \label{fig:icp_example}
\end{figure*}

\subsection{6D Pose Annotation} \label{subsec:pose_annotation}

\begin{figure}[t!]
 \centering
    \includegraphics[width=\linewidth]{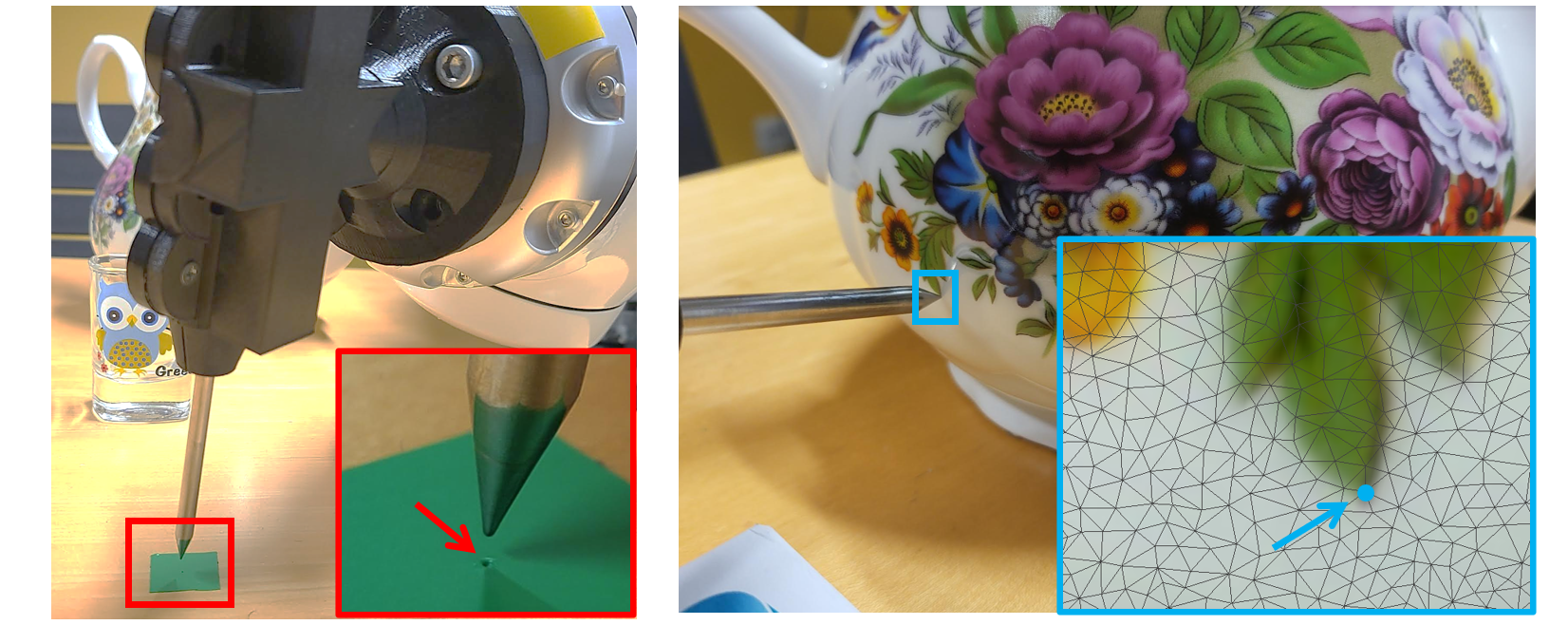}
    \caption{Tip calibration (left) with its pivot point (red). Tip measuring points of object surface (right) and its correspondence on the object's model mesh (blue).}
    \label{fig:tipping}
\end{figure}

Annotating the precise 6D pose of the objects is a challenging task as mentioned in Sec.~\ref{subsec:annotation_problem}. Here, we utilize the robot accuracy and its reproducible encoders to annotate the object pose. Our annotation steps are as follows: first, we attach the tool tip on the robot's end-effector and measure several keypoints along with 20-30 surface points of the given object by hand guiding the end-effector while the robot is in gravity compensation mode (Fig~\ref{fig:pipeline_overview} (c), Fig~\ref{fig:tipping} (right)). Then, corresponding keypoints are manually picked on the object model's mesh to obtain the initial pose of the respective objects (Fig~\ref{fig:tipping} (right) (blue)). Finally, ICP is applied to align the dense mesh points of the object and the measured sparse surface points as the refinement step for the initial object poses.%

To evaluate the refinement performance, 25 points on a specific area of the object surface are picked  and  uniformly distributed noise of \(\pm\)0.2mm is added to simulate the measurement noise. We then apply a small perturbation of random translation errors of range \(\pm\)2mm in x,y,z and a rotation error about a random axis with an angle of up to 4 degrees to the object pose to simulate the error introduced by the point correspondences.  Thereafter, we apply ICP between the picked surface points and the perturbed mesh to refine the pose. We test this pipeline with 3 selected objects with 5 different random perturbation before applying ICP to recover the initial pose. The pose error is measured in translation and rotational distance~\cite{laval_error} after the refinement and it gives an average RMSE of \(0.20\,\text{mm}\) in translation and $0.38^{\circ}$ in rotation.

It is observed that ICP improves the annotation in real life scenario particularly on textureless objects, where it is difficult to find exact correspondence from the mesh. An extreme example of annotated poses before and after ICP on the textureless objects is shown in Fig.~\ref{fig:icp_example}.

\subsection{Hand-Eye Calibration} \label{subsec:hand_eye_calibration}
Traditional hand-eye calibration, such as \cite{tsai1989new} requires detection of the marker from the camera in various positions to obtain an accurate calibration result. The transformation from camera to end-effector is difficult to estimate as the marker transformation to robot base is unknown and both have to be jointly estimated. In our case,  however, the marker position can be accurately measured with the robot tip. Considering this fact, we measure 12 selected points on the marker board and calculate \(T_{Marker\to RB}\) (Fig.~\ref{fig:handeye} (a)) to link the end-effector pose to the camera frame. From \(T_{Marker\to RB}\),  the \(T_{handeye}\) is calculated as shown in Fig.~\ref{fig:handeye} (b). 

\begin{figure}[!ht]
 \centering
    \includegraphics[width=\linewidth]{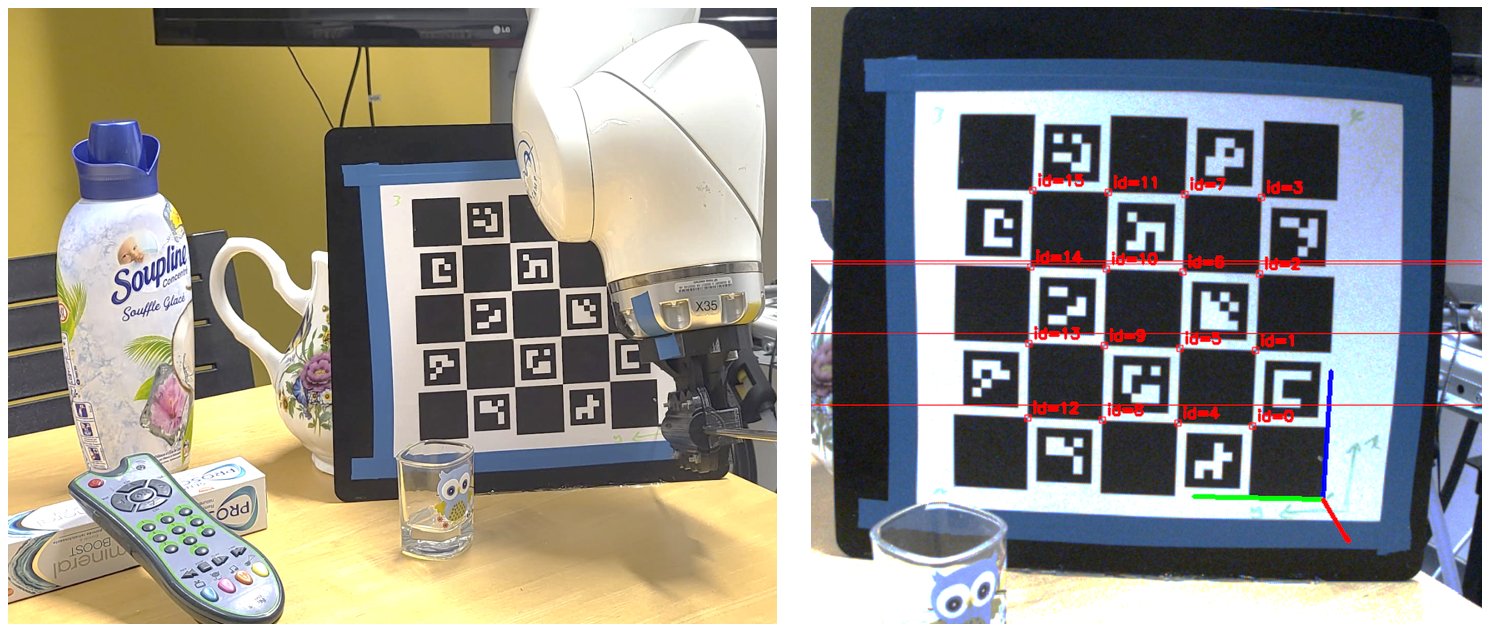}
    \caption{Measuring the marker points for the calibration on the scene (left) and detected marker from one of the cameras (right)}
    \label{fig:hand_eye_real}
\end{figure}

The overall accuracy of the entire procedure is measured as shown in Fig.~\ref{fig:handeye} (c). \(T_{Marker\to cam}\) is formed by applying \(T_{handeye}\) and multiply \(T_{marker\to cam}\) of \(n\) different views to transform the 12 points from the marker board to the robot base (\({P_{transformed}}_{n}\)). RMSE is calculated by comparing the result to \(P_{measured}\). We  evaluate our hand-eye calibrations in one of our scenes on both RGBD and Polarization camera with the mentioned approach with \(n=10\) and obtained \(\text{RMSE}_{\text{RGBD}} = 0.89 \, \text{mm}\) and  \(\text{RMSE}_{\text{Polarization}} =  0.83 \, \text{mm}\)
 across all the view points. This calibration is performed procedure for all cameras before recording each scene as shown in Fig.~\ref{fig:hand_eye_real}.

\subsection{Synchronized Robot Pose Capture with Images}
RGBD and polarization cameras are used for the data acquisition. A specially designed and 3D printed rig is used to mount both cameras tightly on the end-effector. The trajectory of all joints of the robot is recorded by manually moving the end-effector while the robot arm is in gravity compensated mode. Thereafter, we record the images of the scene by replaying the joint trajectory while stopping the robot every 5-7 joint positions to capture the images and the robot pose (approx 10-15 fps). This ensures no motion blur and camera synchronization artefacts are recorded while reproducing the original hand-held camera trajectory.

\subsection{Evaluation of Overall Annotation Quality}

We evaluate overall annotation quality of our dataset by running simulated data acquisition with two measured error statistics : ICP error (Sec~\ref{subsec:pose_annotation}) and hand-eye calibration error (Sec~\ref{subsec:hand_eye_calibration}). For both RGBD and Polarization camera, setup from one of the scenes is used including the objects and the trajectories. The acquisition is simulated twice, with and without the aforementioned error. In the end, RMSE error is calculated pointwise in mm between the acquisitions. We averaged the error per object and per each frame in the trajectories.

RMSE error for RGBD camera is 0.84 mm and for polarization camera is 0.76 mm. Detailed description of this procedure is attached in the supplementary material. The annotation quality in comparison with other dataset acquisition principles is shown in Tab. \ref{tab:pose_accuracy}.

\begin{table}[htb]
    \centering
    \footnotesize
    \resizebox{\columnwidth}{!}{
    \begin{tabular}{l c c c c}
    \toprule
    Dataset & RGBD Dataset & TOD~\cite{liu2020keypose} & StereOBJ~\cite{liu2021stereobj} & Ours\\
    \hline
    3D Labeling & Depth Map & \multicolumn{2}{c}{Multi-View} & Robot\\
    Point RMSE & $\geq 17$mm & $3.4$mm & $2.3$mm & $0.80$mm\\
    \bottomrule
    \end{tabular}
    }
    \caption{Comparison of pose annotation quality for different dataset setups. The error for RGBD is exemplified with the standard deviation of the Microsoft Azure Kinect~\cite{liu2021stereobj}.}
    \label{tab:pose_accuracy}
\end{table}

%% file: sections/08_experiments.tex
\begin{table*}[t!]
    \centering
    \resizebox{\textwidth}{!}{
    \begin{tabular}{|lccccccccc|}
        \hline
        3D$_{25}$ / 3D$_{50}$  & Bottle & Box & Can & Cup & Remote & Teapot & Cutlery & Glassware & Mean\\
        \hline
        NOCS \cite{wang2019normalized} & \textbf{91.17} / 0.65 & 16.10 / 0.01 & \textbf{85.44} / \textbf{23.01} & 51.83 / 1.48 & \textbf{93.26} / \textbf{86.05} & 0.00 / 0.00 & 4.89 / 0.01 & 4.00 / 0.06 &  43.34 / 13.91\\
        \hline
        CPS \cite{manhardt2020cps} & 80.08 / \textbf{40.30}  & \textbf{31.68} / \textbf{28.18}  & 68.96 / 6.69  & \textbf{81.60} / \textbf{70.24} & 86.30 / 37.08  & \textbf{67.43} / \textbf{4.31} & \textbf{44.00} / \textbf{24.95}  & \textbf{30.33} / \textbf{17.74} & \textbf{61.30} / \textbf{28.69} \\ 
        \hline
    \end{tabular}}
    \caption{Class-wise evaluation of 3D IoU for NOCS~\cite{wang2019normalized} and CPS~\cite{manhardt2020cps} on test split of known objects.}
    \label{tab:test_result_1}
\end{table*}

\begin{table*}[t!]
    \centering
    \resizebox{\textwidth}{!}{
    \begin{tabular}{|lccccccccc|}
        \hline
        3D$_{25}$ / 3D$_{50}$  & Bottle & Box & Can & Cup & Remote & Teapot & Cutlery & Glassware & Mean\\
        \hline
        Experiment 1 & \textbf{91.17} / 0.65 & 16.10 / \textbf{0.01} & \textbf{85.44} / \textbf{23.01} & 51.83 / \textbf{1.48} & \textbf{93.26} / \textbf{86.05} & 0.00 / 0.00 & 4.89 / \textbf{0.01} & \textbf{4.00} / \textbf{0.06} &  \textbf{43.3} / \textbf{13.91}\\
        \hline
        Experiment 2 & 13.70 / \textbf{1.28}  & \textbf{27.74} / 0.00  & 48.17 / 0.00  & \textbf{61.77} / 0.00 & 8.35 / 0.00  & \textbf{4.90} / 0.00 & \textbf{16.10} / 0.00  & 0.83 / 0.00 & 22.70 / 0.17 \\ 
        \hline
    \end{tabular}}
    \caption{Class-wise evaluation of 3D IoU for NOCS~\cite{wang2019normalized} on seen (Experiment 1) and mostly unseen (Experiment 2) objects.}
    \label{tab:test_result_2}
\end{table*}

\begin{figure*} [t!]
     \centering
     \begin{subfigure}[b]{0.33\textwidth}
         \centering
         \includegraphics[width=\textwidth]{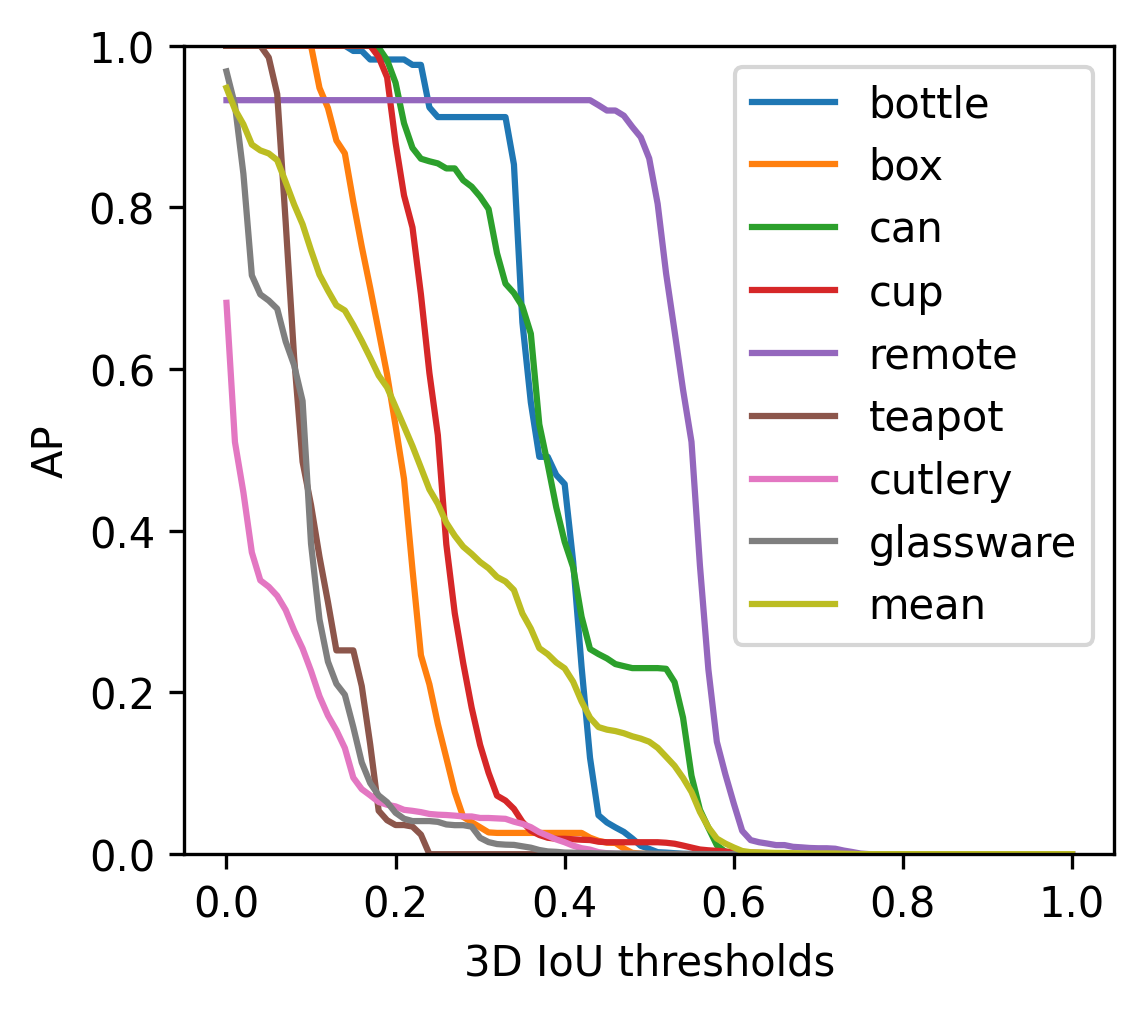}
         \caption{NOCS result in the first experiment }
         \label{fig:plot_1}
     \end{subfigure}
     \hfill
     \begin{subfigure}[b]{0.33\textwidth}
         \centering
         \includegraphics[width=\textwidth]{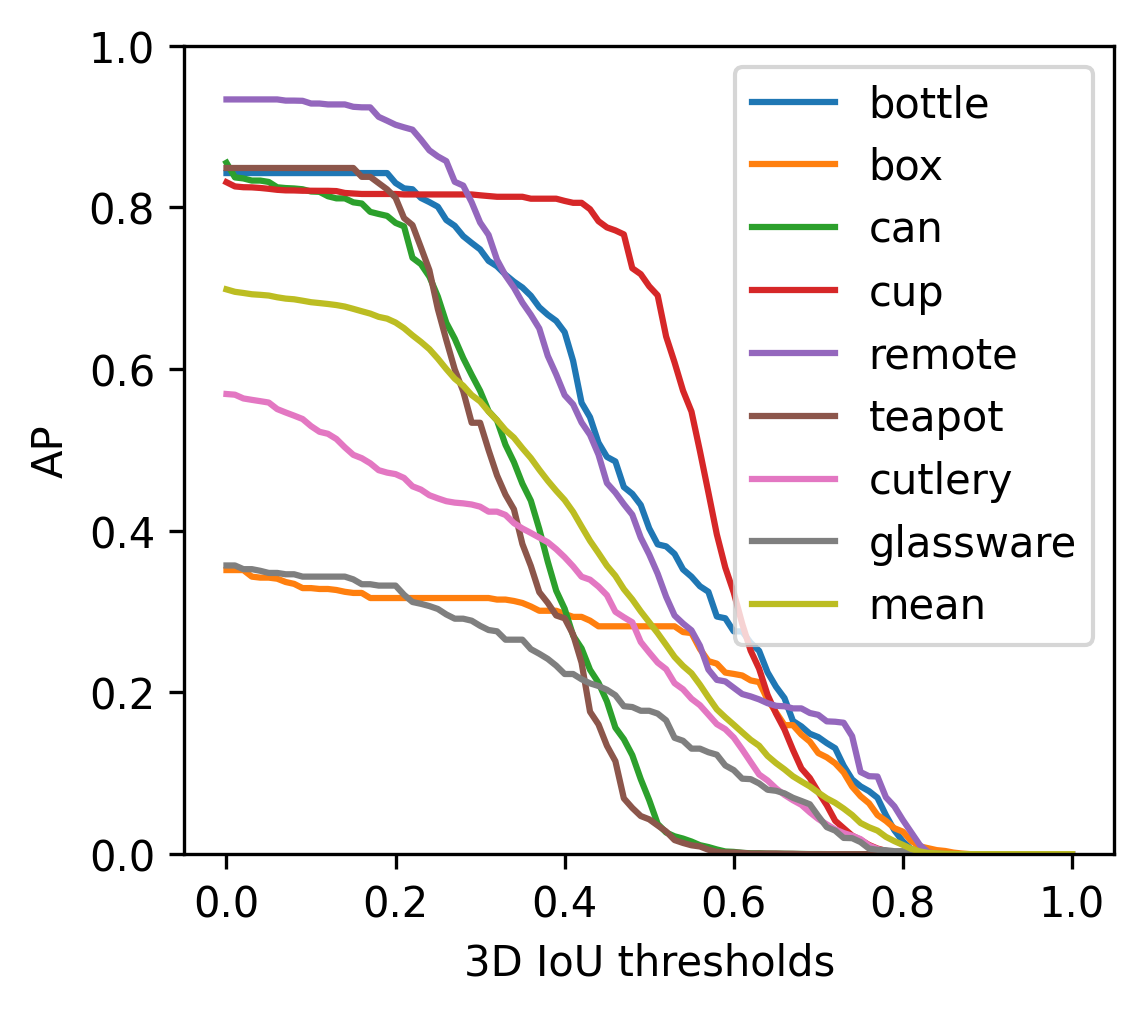}
         \caption{CPS result in the first experiment }
         \label{fig:plot_2}
     \end{subfigure}
     \hfill
     \begin{subfigure}[b]{0.33\textwidth}
         \centering
         \includegraphics[width=\textwidth]{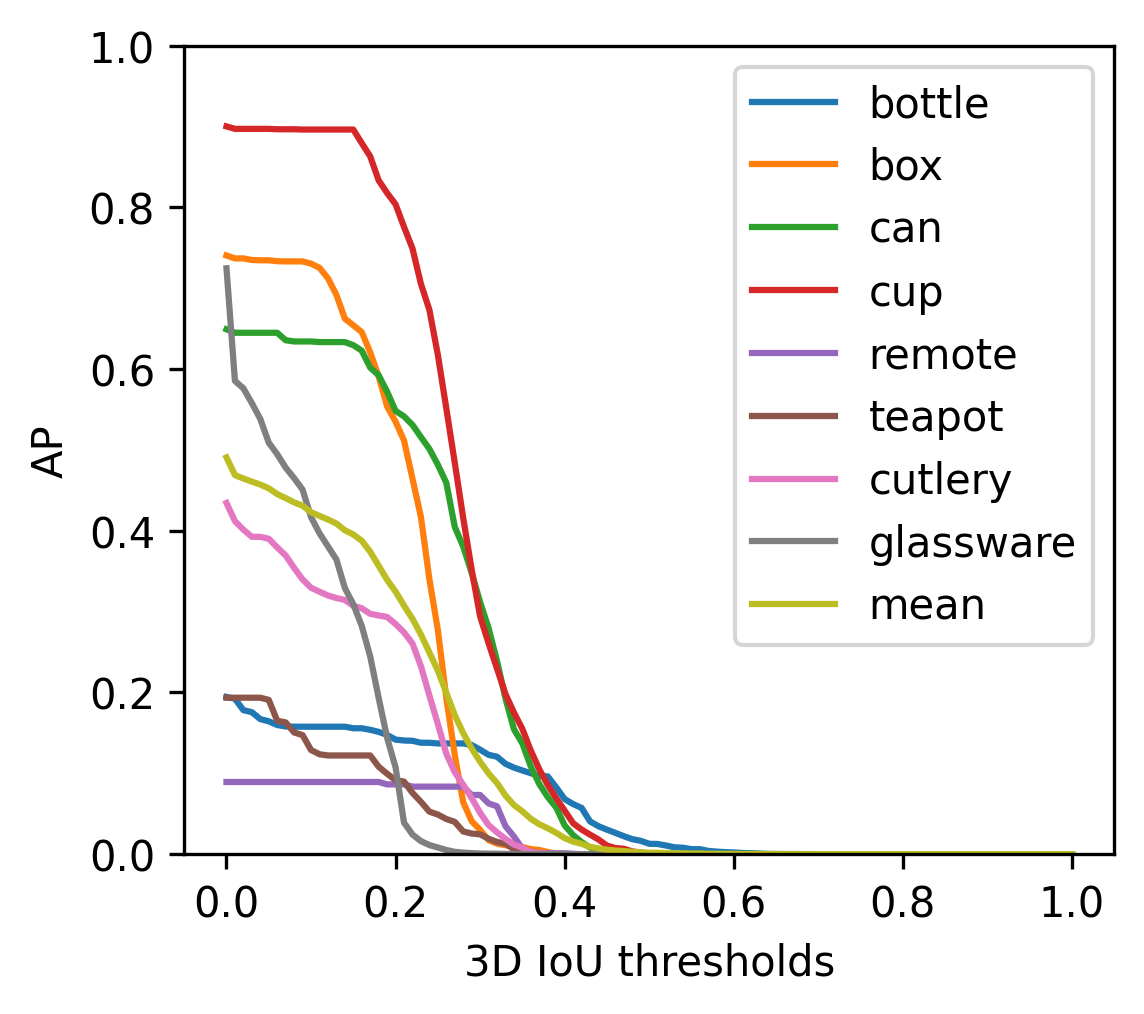}
         \caption{NOCS result in the second experiment }
         \label{fig:plot_3}
     \end{subfigure}
        \caption{Plots of average precision (AP) with respect to 3D IoU thresholds for each category. }
        \label{fig:experiment_plots}
\end{figure*}

\section{Benchmarks and Experiments}

Both monocular (CPS) and RGB-D based (NOCS) category-level methods are considered for the baseline evaluation on the PhoCaL dataset. For the evaluation of NOCS, the normal object coordinate space maps are rendered for each training image and will be published together with the dataset. With the predicted normalized object shape from NOCS map, the depth information is used to lift 2D detection to 3D space using ICP. Considering the artifacts in the depth data from metallic and transparent objects in the dataset, along with the occlusion, the test sequences are very challenging for RGBD methods. 

Similiar to NOCS, CPS first detects 2D bounding boxes. Then lifting modules for each class transform 2D image features to 6D pose and scales. Simultaneously the method also estimates the point cloud shape for the respective object class. CPS is trained on approximately 1000 object instance models for each category to learn a deep point cloud encoding of each class. The 2D detection and lifting modules are trained together for 100k steps with a learning rate of 1e-4, decaying to 1e-5 at 60k steps.

\subsection{Evaluation Pipeline}
Our dataset consists of 24 image sequences in total with training and testing split in each sequence. In our evaluation pipeline, the training split of the first 12 sequences are used to train the network. To have an evaluation on both the known and novel objects in each category, two experiments are designed. To evaluate on seen objects firstly, the network is trained on the training split of the first 12 sequences and tested on the testing split of the same sequences. To further evaluate the generalization ability of NOCS and CPS to novel objects in the same category, the same training split of the first 12 sequences is used, but we evaluate the result on the testing split of the latter 12 sequences, where objects are mostly unseen. With this way, generalization ability of the methods to novel objects in the category is emphasized, which is a  common issue in real operating environments. The evaluation metric is the intersection over union (IoU) result with a threshold of 25\% and 50\%.

\subsection{Evaluation Result}
The 3D IoU at 25\% and 50\% evaluations of NOCS for the first experiment setup is shown in  Tab. \ref{tab:test_result_1}. The mean average precision (mAP) for 3D IoU at 25\% is 43.34\%. It is observed in the experiment that even if the segmentation and normalized object coordinate map predictions are accurate, the lifting from NOCS map to 6D space is sensitive to artifacts in depth maps. Since the objects are highly occluded in the PhoCaL dataset, and depth measurements are inaccurate because of cutlery and glassware categories, the method does not have a good performance on the dataset which indicates the drawbacks of RGBD methods in these photometrically challenging cases.
The average precision of each category with respect to 3D IoU threshold is plotted in Fig. \ref{fig:plot_1}. Note that the results of cutlery and glassware categories are among the worst three categories. 

For comparison, the result of CPS is also listed in Tab. \ref{tab:test_result_1}. As can be seen from the table, CPS has a higher precision for cutlery and glassware categories. Monocular methods are not affected by artifacts in depth images, which explains the result from the experiment. CPS has a higher mAP of 61.30\%, which means RGB has an advantage in dealing with photometrically challenging objects. The detailed APs for each category are plotted in Fig. \ref{fig:plot_2}.

In addition, the NOCS evaluation on both experiments are compared in table \ref{tab:test_result_2}. The evaluation result for the second experiment has a lower mAP for 3D IoU at 25\% and 50\% as expected, as most of the test objects are novel in the second experiment. Fig. \ref{fig:plot_3} plots NOCS APs in the second experiment. In comparison to NOCS, the CPS result drops significantly in the second experiment and the 3D IoU  at 25\% is 4.3\%. The result shows that pretraining with a large amount of synthetics images is necessary for monocular methods, to learn the correct lifting from 2D detection to 3D space without the help of depth images.

%% file: sections/10_conclusion.tex
\subsection{Limitations}
Even though the proposed pipeline for annotating the 6D pose ground truth is accurate, annotating the objects with deformable surface, such as empty boxes, poses a challenge during the surface measurement step in the workflow due to its light deformation which could deteriorate the quality of both initial pose and ICP based refinement. Moreover, the limited workspace of the robot constrains the view angles in the image sequences which is an issue the PhoCaL shares with other robotic acquisition setups.
The hand eye calibration of the camera plays a key role for the annotation quality. If the camera resolution is low, a good calibration result requires significantly more input images from different angles.     %

\section{Conclusion}
In this paper we introduce the PhoCaL dataset, which contains photometrically challenging categories. High-quality 6D pose annotations are provided  for all categories and multiple camera modalities, namely RGBD and RGBP. With our manipulator-driven annotation pipeline, we reach pose accuracy levels that are one order of magnitude more precise than previous vision-sensor-only pipelines even for photometrically complex objects. Moreover, baselines are provided for future works on category-level 6D pose on our dataset by evaluating both monocular and RGB-D methods. The evaluation shows the difficulty level of the dataset in particular for objects that include reflective and transparent surfaces. PhoCaL therefore constitutes a challenging dataset with accurate ground truth that can pave the way for future pose pipelines that are applicable to more realistic scenarios with everyday objects.